\documentclass{article}
\pdfoutput=1 
 \PassOptionsToPackage{numbers, compress}{natbib}
\usepackage[final]{nips_2016}

\usepackage[utf8]{inputenc} 
\usepackage[T1]{fontenc}    
\usepackage{hyperref}       
\usepackage{url}            
\usepackage{booktabs}       
\usepackage{amsfonts}       
\usepackage{nicefrac}       
\usepackage{microtype}      
\usepackage{graphicx}
\usepackage{amsmath}
\usepackage{amssymb}
\usepackage{subfigure}
\usepackage{epsfig}
\usepackage{verbatim}
\usepackage{multirow}
\usepackage{color}
\usepackage{algorithm}
\usepackage{algpseudocode}

\title{Can Peripheral Representations Improve Clutter Metrics on Complex Scenes?}

\author{
  Arturo Deza \\
  Dynamical Neuroscience\\
  Institute for Collaborative Biotechnologies \\
  UC Santa Barbara, CA, USA\\
  \texttt{deza@dyns.ucsb.edu} \\
   \And
   Miguel P. Eckstein \\
   Psychological and Brain Sciences \\
   Institute for Collaborative Biotechnologies \\
   UC Santa Barbara, CA, USA \\
   \texttt{eckstein@psych.ucsb.edu} \\
}

\begin{document}

\maketitle
\vspace{-10pt}
\begin{abstract}
Previous studies have proposed image-based clutter measures that correlate with human search times
and/or eye movements. However, most models do not take into account the fact that the effects of
clutter interact with the foveated nature of the human visual system: visual clutter further from the fovea
has an increasing detrimental influence on perception. Here, we introduce a new foveated clutter model
to predict the detrimental effects in target search utilizing a forced fixation search task. We use Feature
Congestion (Rosenholtz et al.) as our non foveated clutter model, and we stack a peripheral architecture
on top of Feature Congestion for our foveated model. We introduce the Peripheral Integration Feature
Congestion (PIFC) coefficient, as a fundamental ingredient of our model that modulates clutter as a non-linear 
gain contingent on eccentricity. We finally show that Foveated Feature Congestion (FFC) clutter
scores $(r(44)=-0.82\pm0.04,p<0.0001)$ correlate better with target detection (hit rate) than regular
Feature Congestion $(r(44)=-0.19\pm0.13,p=0.0774)$ in forced fixation search. Thus, our model allows us to enrich
clutter perception research by computing fixation specific clutter maps. A toolbox for creating peripheral
architectures: \textit{Piranhas}: \textbf{P}er\textbf{i}phe\textbf{r}al \textbf{A}rchitectures for \textbf{N}atural, \textbf{H}ybrid 
and \textbf{A}rtificial \textbf{S}ystems will be made available\footnote{\url{https://github.com/ArturoDeza/Piranhas}}.
\end{abstract}

\section{Introduction}
\vspace{-10pt}
What is clutter? While it seems easy to make sense of a cluttered desk \textit{vs} an uncluttered desk at a
glance, it is hard to quantify clutter with a number. Is a cluttered desk, one stacked with papers?
Or is an uncluttered desk, one that is more organized irrelevant of number of items? An important goal in
clutter research has been to develop an image based computational model that outputs a quantitative
measure that correlates with human perceptual behavior~\cite{oliva2004identifying,henderson2009influence,rosenholtz2007measuring}. 
Previous studies have created models that output global or regional metrics to measure clutter perception. Such measures are aimed to
predict the influence of clutter on perception. However, one important aspect of human visual perception
is that it is not space invariant: the fovea processes visual information with high spatial detail while
regions away from the central fovea have access to lower spatial detail. Thus, the influence of clutter on
perception can depend on the retinal location of the stimulus and such influences will likely interact with
the information content in the stimulus. 

The goal of the current paper is to develop a foveated clutter
model that can successfully predict the interaction between retinal eccentricity and image content in
modulating the influence of clutter on perceptual behavior. We introduce a foveated mechanism
based on the peripheral architecture proposed by Freeman and Simoncelli~\cite{freeman2011metamers} and stack it into a current clutter
model (Feature Congestion~\cite{rosenholtz2005feature,rosenholtz2007measuring}) to generate a clutter map that arises from a calculation of
information loss with retinal eccentricity but is multiplicatively modulated by the original unfoveated clutter
score. The new measure is evaluated in a gaze-contingent psychophysical experiment measuring target
detection in complex scenes as a function of target retinal eccentricity. We show that the foveated clutter
models that account for loss of information in the periphery correlates better with human target detection
(hit rate) across retinal eccentricities than non-foveated models. Although the model is presented in the
context of Feature Congestion, the framework can be extended to any previous or future clutter metrics
that produce clutter scores that are computed from a global pixel-wise clutter map.

\section{Previous Work}
\vspace{-10pt}
Previous studies have developed general measures of clutter computed for an entire image and do not
consider the space-variant properties of the human visual system. Because our work seeks to model and
assess the interaction between clutter and retinal location, experiments manipulating the eccentricity of a
target while observers hold fixation (gaze contingent forced fixation) are most appropriate to evaluate the
model. To our knowledge there has been no systematic evaluation of fixation dependent clutter models
with forced fixation target detection in scenes.
In this section, we will give an overview of state-of-the-art clutter models, metrics and evaluations.

\subsection{Clutter Models}

\begin{figure}[t]
\centering
\includegraphics[scale=0.13,clip=true,draft=false,]{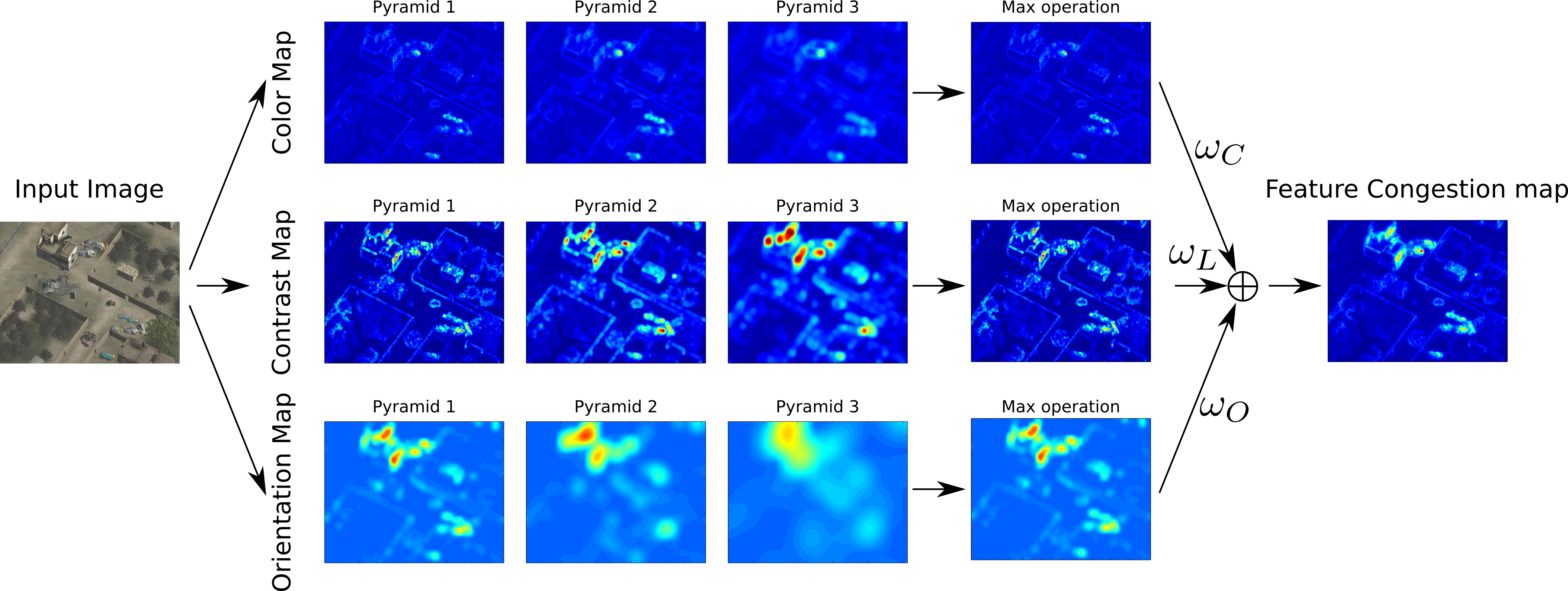}
\vspace{-10pt}
\caption{The Feature Congestion pipeline as explained in Rosenholtz \textit{et al.}~\cite{rosenholtz2007measuring}. A color, contrast and orientation feature map 
for each spatial pyramid is extracted, and the max value of each is computed as the final feature map.
The Feature Congestion map is then computed by a weighted sum over each feature map. The Feature Congestion score is the mean value of the map.
}\label{fig:FC_pipeline}
\end{figure}

\textbf{Feature Congestion:} Feature Congestion, initially proposed by~\cite{rosenholtz2005feature,rosenholtz2007measuring} produces both a pixel-wise clutter score map as a well as a global clutter
score for any input image or Region of Interest (ROI). Each clutter map is computed by combining a Color map in CIELab space,
an orientation map~\cite{landy1991texture}, and a local contrast 
map at multiple scales through Gaussian Pyramids~\cite{burt1983laplacian}.
One of the main advantages Feature Congestion has is that each pixel-wise clutter score (Fig.~\ref{fig:FC_pipeline}) and global score can be computed in less than a second. Furthermore,
this is one of the few models that can output a specific clutter score for any pixel or ROI in an image. 
This will be crucial for developing a foveated model as explained in Section~\ref{Sec:FFC}.\\
\textbf{Edge Density:} Edge Density computes a ratio after applying an Edge Detector on the input image~\cite{oliva2004identifying}. The final clutter score is the ratio of edges to total number of pixels
present in the image. The intuition for this metric is straightforward: ``the more edges, the more clutter'' (due to objects for example).\\
\textbf{Subband Entropy:} The Subband Entropy model begins by computing $N$ steerable pyramids~\cite{simoncelli1995steerable} at $K$ orientations across each channel from the 
input image in CIELab color space. 
Once each $N\times K$ subband is collected for each channel, 
the entropy for each oriented pyramid is computed pixelwise and they are averaged separately. 
Thus, Subband Entropy wishes 
to measure the entropy of each spatial frequency and oriented filter response of an image.\\
\textbf{Scale Invariance}: The Scale Invariant Clutter Model proposed by Farid and Bravo~\cite{bravo2008scale} uses graph-based 
segmentation~\cite{felzenszwalb2004efficient} at multiple $k$ scales. A scale invariant clutter representation is given by the power law 
coefficient that matches the decay of number of regions with the adjusted scale parameter.\\
\textbf{ProtoObject Segmentation:} ProtoObject Segmentation proposes an unsupervised metric for clutter scoring~\cite{yu2013modeling,yu2014modeling}. The model begins by converting the image
into HSV color space, and then proceeds to segment the image through superpixel segmentation~\cite{liu2011entropy,levinshtein2009turbopixels,achanta2010slic}. 
After segmentation, mean-shift~\cite{fukunaga1975estimation} is applied on all cluster (superpixel) medians to calculate the final amount of representative colors present in the image. 
Next, superpixels are merged with one another contingent on them being adjacent,
and being assigned to the same mean-shift HSV cluster. The final score is a ratio between initial number of superpixels and final number of superpixels. \\
\textbf{Crowding Model:} The Crowding Model developed by van der Berg~\textit{et al.}~\cite{van2009crowding} is the only model to have used losses in the periphery due to 
crowding as a clutter metric. It decomposes the image into 3 different scales in CIELab color space. It then produces 6 different orientation maps for each 
scale given the luminance
channel; a contrast map is also obtained by difference of Gaussians on the previously mentioned channel. All feature maps are then pooled with 
Gaussian kernels that grow linearly
with eccentricity, KL-divergence is then computed between the pre and post pooling feature maps to get information loss coefficients, 
all coefficients are averaged together to produce a final clutter score.
We will discuss the differences of this model to ours in the Discussion (Section~\ref{Sec:Discussion}).\\
\textbf{Texture Tiling Model:} The Texture Tiling Model (TTM) is a recent perceptual model that accounts for 
losses in the periphery~\cite{rosenholtz2012summary,keshvari2016pooling} through psyhophysical experiments modelling visual search~\cite{eckstein2011visual}: feature search, conjunction search,
configuration search and asymmetric search. In essence, the Mongrels
proposed by Rosenholtz \textit{et al.} that simulate peripheral losses 
are very similar to the Metamers proposed by Freeman \& Simoncelli~\cite{freeman2011metamers}.
We do not include comparisons to the TTM model since it requires additional psychophysics on the Mongrel versions of the images.

\begin{figure}[t]
\centering
\includegraphics[scale=0.12,clip=true,draft=false,]{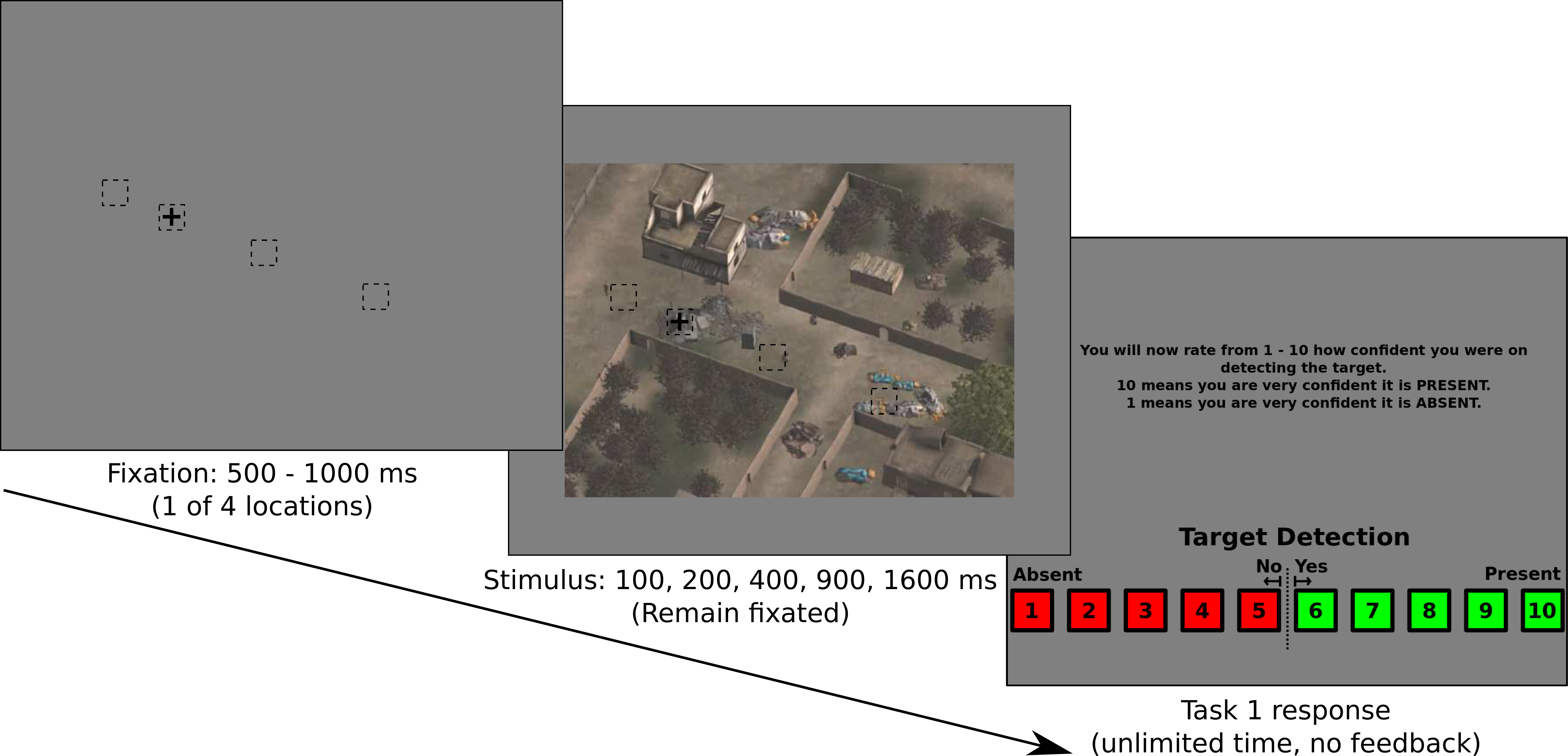}
\caption{Experiment 1: Forced Fixation Search flow diagram. A naive observer begins by fixating the image at a location that is either 
1, 4, 9 or 15 $\deg$ away from the target (the observer is not aware of the possible eccentricities). 
After fixating on the image for a variable amount of 
time (100, 200, 400, 900 or 1600 ms), the observer
must make a decision on target detection.
}\label{fig:Experiment_Flow}
\vspace{-10pt}
\end{figure}

\subsection{Clutter Metrics}
\vspace{-5pt}
\textbf{Global Clutter Score:} The most basic clutter metric used in clutter research is the original clutter score that every model computes over the entire image. Edge Density \& 
Proto-Object Segmentation output a ratio, while Subband Entropy and Feature Congestion output a score. However, Feature Congestion is the only model that
outputs a dense pixelwise clutter map before computing a global score (Fig.~\ref{fig:FC_pipeline}). Thus, we use Feature Congestion clutter maps for our foveated clutter model.\\
\textbf{Clutter ROI:} The second most used clutter metric is ROI (Region of Interest)-based, as shown in the work of Asher \textit{et al.}~\cite{asher2013regional}. 
This metric is of interest when an observer is engaging in target search, \textit{vs} making a human judgement (Ex: ``rate the clutter of the 
following scenes'').

\subsection{Clutter Evaluations}
\vspace{-5pt}
\textbf{Human Clutter Judgements:} Multiple studies of clutter, correlate their metrics with rankings/ratings of clutter provided by human participants. Ideally, if clutter 
model A is better than clutter model B, then the correlation of model scores and human rankings/ratings should be higher for model A than for model B.~\cite{yu2014modeling,oliva2004identifying,van2009crowding}\\
\textbf{Response Time:} Highly cluttered images will require more time for target search, hence more time to arrive to a decision of target present/absent. 
Under the previous assumption, a high correlation value between response time and clutter score are a good sign for a clutter model.~\cite{rosenholtz2007measuring,bravo2008scale,van2009crowding,asher2013regional,henderson2009influence}\\
\textbf{Target Detection (Hit Rate, False Alarms, Performance):} In general, when engaging in target search for a fixed amount of time across all trial conditions, 
an observer will have a lower hit rate and higher false alarm rate for a highly cluttered image than an uncluttered image.~\cite{rosenholtz2007measuring,asher2013regional,henderson2009influence}

\section{Methods \& Experiments}
\vspace{-10pt}

\subsection{Experiment 1: Forced Fixation Search}
\label{Sec:FF_Search}

A total of 13 subjects participated in a Forced Fixation Search experiment where the goal was to detect a target in the subject's periphery 
and identify if there was a target (person) present or absent. 
Participants had variable amounts
of time (100, 200, 400, 900, 1600 ms) to view each clip that was presented in a random order at a variable degree of eccentricities that the subjects were 
not aware of ($1\deg$, $4\deg$, $9\deg$, $15\deg$). 
They were then prompted with a Target Detection rating scale
where they had to rate from a scale from 1-10 by clicking on a number reporting how confident they were on detecting the target. 
Participants have unlimited time for making their judgements, 
and they did not take more than 10 seconds per judgment. There was no response feedback after each trial.
Trials were aborted when subjects broke fixation outside of a $1\deg$ radius around the fixation cross.

Each subject did 12 sessions that consisted of 360 unique images. Every session also presented the images with aerial viewpoints from different vantage 
points (Example: session 1 had the target at 12 o'clock, while session 2 had the target at 3 o'clock). To control for any fixational biases, all subjects had a unique
fixation point for every trial for the same eccentricity values. All images were rendered with variable levels of clutter. Each session took about an hour to complete.
The target was of size $0.5\deg\times 0.5\deg$, $1\deg\times 1\deg$, $1.5\deg\times 1.5\deg$, depending on zoom level. 

For our analysis, we only used the low zoom and 100 ms time condition since there was less ceiling effects across all eccentricities. 

\textbf{Stimuli Creation:} A total of 273 videos were created each with a total duration of 120 seconds, where a `birds eye' point-of-view camera 
rotated slowly around the center. While the video was in rotating motion, there was no relative motion between any parts of the video. 
From the original videos, a total of $360\times 4$ different clips were created. Half of the clips were target present, while the other half were target absent.
These short and slowly rotating clips were used instead of still images in our experiment, to simulate slow real movement from a pilot point of view.
All clips were shown to participants in random order.

\textbf{Apparatus:} An EyeLink 1000 system (SR Research) was used to collect Eye Tracking data at a frequency of 1000Hz. Each participant was at a distance of 76 cm from a LCD screen
on gamma display, so that each pixel subtended a visual angle of $0.022\deg/px$. All video clips were rendered at $1024\times 760$ pixels $(22.5\deg\times 16.7\deg)$ and a frame rate of 24fps.
Eye movements with velocity over $22\deg/s$ and acceleration over $4000\deg/s^2$ were qualified as saccades. Every trial began with a fixation cross, where
each subject had to fixate the cross with a tolerance of $1\deg$.

\section{Foveated Feature Congestion}
\label{Sec:FFC}
\vspace{-10pt}
A regular Feature Congestion clutter score is computed by taking the mean of the Feature Congestion map of the image or of a target ROI~\cite{henderson2009influence}.
We propose a Foveated Feature Congestion (FFC) model that outputs a score which takes into account two main terms: 1) a regular Feature Congestion (FC) score and 2) a Peripheral Integration 
Feature 
Congestion (PIFC) coefficient that accounts 
the lower spatial resolution of the visual
periphery that are detrimental for target detection. The first term is independent of 
fixation, while the second term will act as a non-linear gain that will either reduce or amplify the clutter score depending on fixation distance from the target. 

In this Section we will explain how to compute a PIFC, which will require creating a human-like peripheral architecture 
as explained in Section~\ref{Sec:Peripheral_Architecture}. We then present our Foveated Feature Congestion (FFC) clutter model in Section ~\ref{Sec:FFC_metric}. 
Finally, we conclude by making a quantiative evaluation of the FFC (Section~\ref{Sec:FFC_evaluation}) in its ability to predict variations of target 
detectability across images and retinal eccentricity of the target.

\begin{figure}[t]
\centering
\subfigure[Top: $g_n(e)$ function. Bottom: $h_n(\theta)$ function.]{
    \includegraphics[scale=0.205,clip=true,draft=false,]{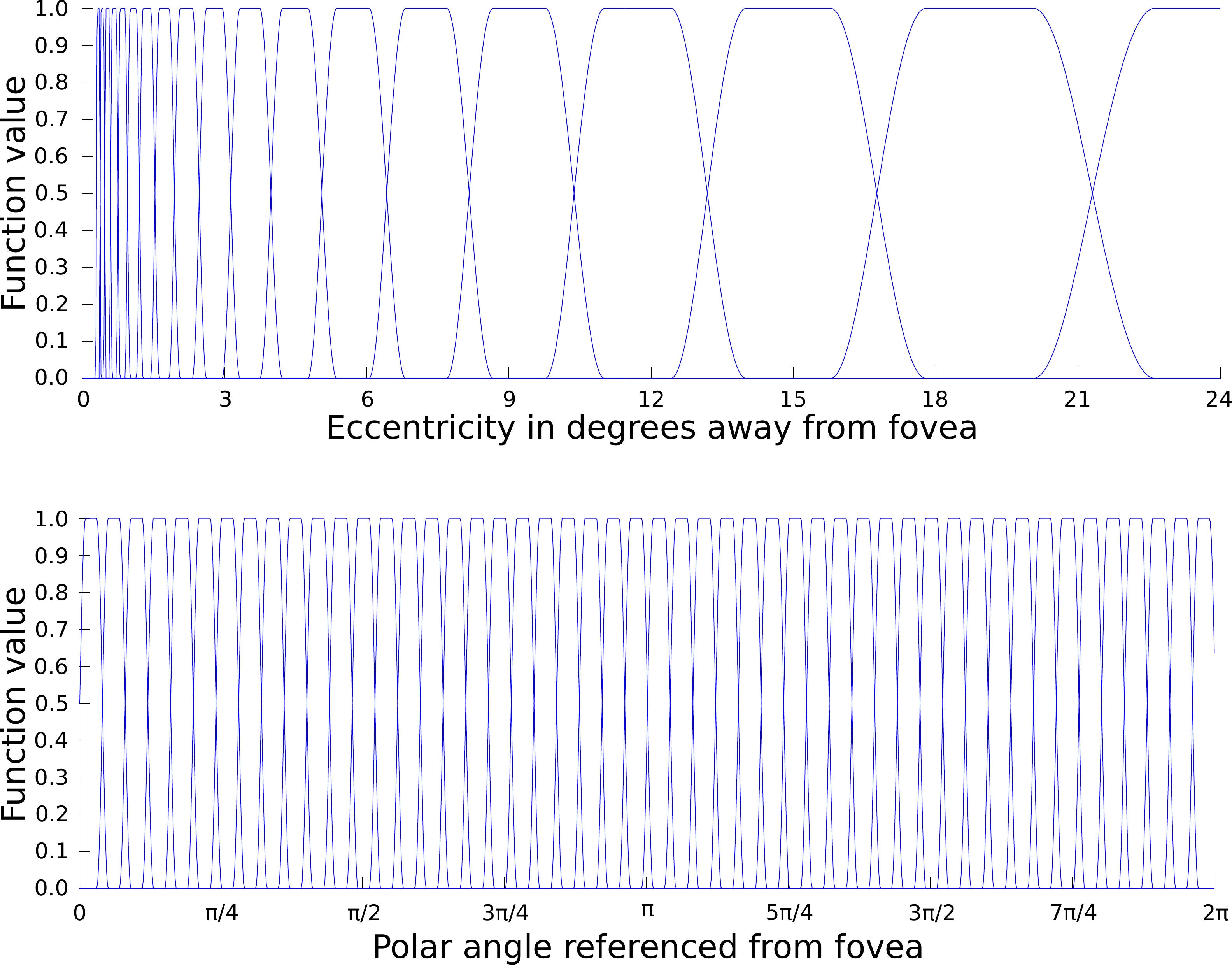}
    \label{fig:FS_logpolar_functions}
}
\subfigure[Peripheral Architecture.]{
    \includegraphics[scale=0.205,clip=true,draft=false,]{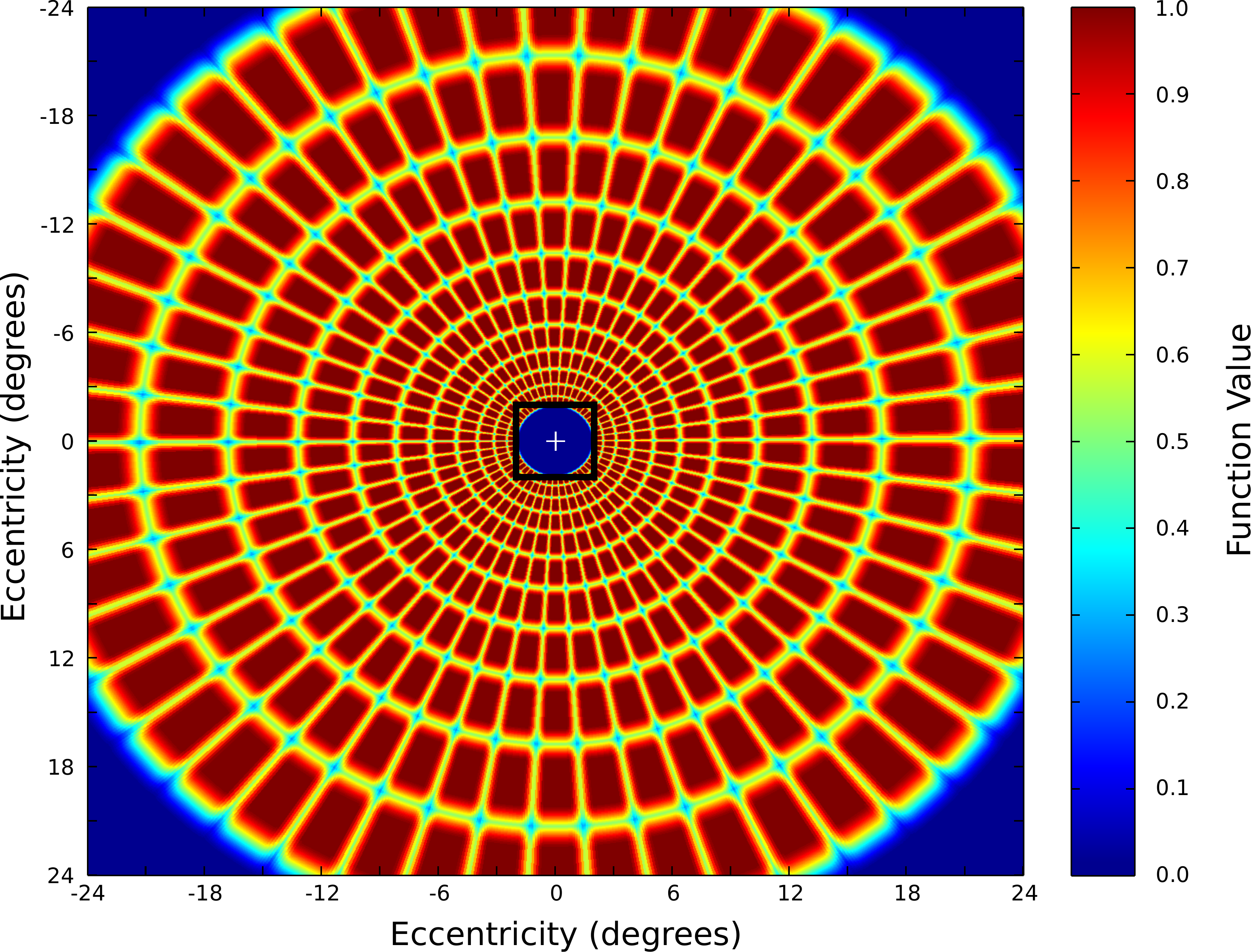}
    \label{fig:peripheral_architecture}
}
\vspace{-10pt}
\caption[]{Construction of a Peripheral Architecture \textit{a la} Freeman \& Simoncelli~\cite{freeman2011metamers} using the functions 
described in Section~\ref{Sec:Peripheral_Architecture} are shown in Fig.~\ref{fig:FS_logpolar_functions}. The blue region in the center of Fig.~\ref{fig:peripheral_architecture}, 
represents the fovea where all information is preserved. Outer regions (in red), represent different parts of the periphery at multiple 
eccentricities.\vspace{-15pt}
}
\label{fig:Pirania}
\end{figure}

\subsection{Creating a Peripheral Architecture}
\label{Sec:Peripheral_Architecture}
\vspace{-5pt}
We used the \textit{Piranhas} Toolkit~\cite{Deza2016piranhas} to create a  Freeman and Simoncelli~\cite{freeman2011metamers} peripheral architecture. 
This biologically inspired model has been
tested and used to model V1 and V2 responses in human and non-human primates with high 
precision for a variety of tasks~\cite{portilla2000parametric,freeman2013functional,movshon2014representation,akbas2014object}. 
It is described by a set of pooling (linear) regions that increase in size with retinal eccentricity.
Each pooling region is separable with respect to polar angle $h_n(\theta)$ and log eccentricity $g_n(e)$,
as described in Eq.~\ref{mod:h} and Eq.~\ref{mod:g} respectively. These functions are multiplied 
for every angle and eccentricity $(\theta,e)$ and are plotted in log polar coordinates to create the peripheral architecture as seen in Fig.~\ref{fig:Pirania}.

\begin{equation}
\label{mod:f}
f(x) = \begin{cases}
        cos^2(\frac{\pi}{2}(\frac{x-(t_0-1)/2}{t_0})); & (-1+t_0)/2<x\leq (t_0-1)/2 \\
        1                                        ; & (t_0-1)/2 < x \leq (1-t_0)/2 \\
        -cos^2(\frac{\pi}{2}(\frac{x-(1+t_0)/2}{t_0}))+1; & (1-t_0)/2 < x \leq (1+t_0)/2
       \end{cases}
\end{equation}

\begin{equation}
\label{mod:h}
 h_n(\theta) = f\Big(\frac{\theta-(w_{\theta}n+\frac{w_{\theta}}{2})}{w_{\theta}}\Big); w_{\theta} = \frac{2\pi}{N_\theta}; n=0,...,N_\theta-1
\end{equation}

\begin{equation}
\label{mod:g}
 g_n(e) = f\Big(\frac{\log(e)-[\log(e_0)+w_e(n+1)]}{w_e}\Big); w_e = \frac{\log(e_r)-\log(e_0)}{N_e}; n=0,...,N_e-1 
\end{equation}

The parameters we used match a V1
architecture with a scale of $s=0.25$ , a visual radius of $e_r = 24\deg$, a fovea of $2\deg$, with $e_0 = 0.25\deg$~\footnote{We remove
regions with a radius smaller than the foveal radius, since there is no pooling in the fovea.}, and $t_0 = 1/2$. The scale defines the number of eccentricities $N_e$, as well as the number of polar pooling regions $N_{\theta}$ from
$\langle 0, 2\pi]$.

Although observers saw the original stimuli at $0.022\deg$/pixel, with image size $1024\times 760$; for modelling purposes: we 
rescaled all images to half their size so the peripheral architecture could fit all images under any fixation point. To preserve stimuli size in degrees
after rescaling our images, our foveal model used an input value of $0.044\deg$/pixel (twice the value of experimental settings).
Resizing the image to half its size also allows the peripheral architecture to consume less CPU computation time and memory.

\begin{figure}[t]
\centering
\includegraphics[scale=0.18,clip=true,draft=false,]{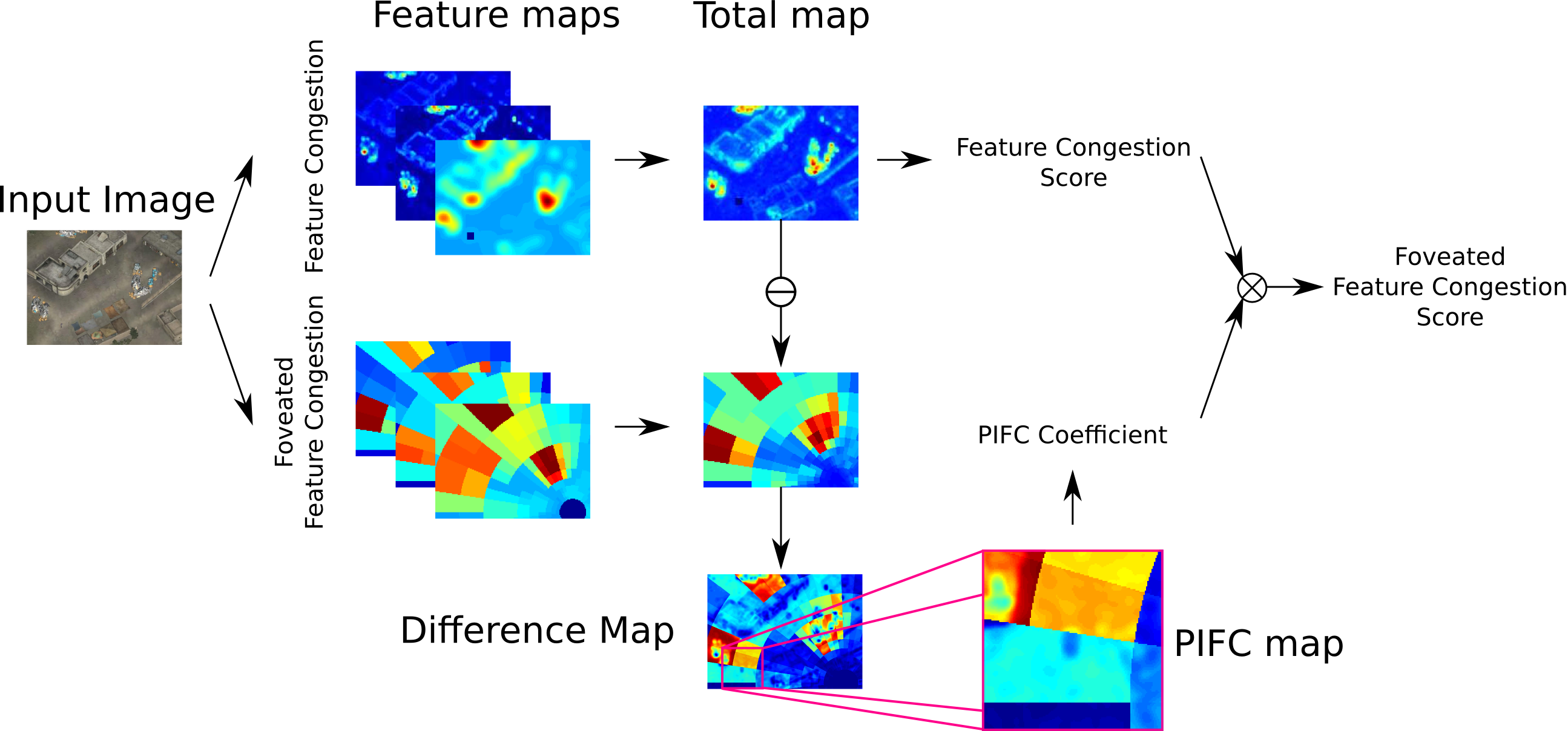}
\vspace{-10pt}
\caption{Foveated Feature Congestion flow diagram: 
In this example, the point of fixation is at $15\deg$  away from the target (bottom right corner of the input image).
A Feature Congestion map of the image (top flow), and a Foveated Feature Congestion map (bottom flow) are created.
The PIFC coefficient is computed around an ROI centered at the target (bottom flow; zoomed box). 
The Feature Congestion score is then multiplied by the PIFC coefficient, and the Foveated Feature Congestion score is returned. 
Sample PIFC's across eccentricities can be seen in the Supplementary Material.
}\label{fig:Foveated_FC}
\end{figure}


\subsection{Creating a Foveated Feature Congestion Model}
\label{Sec:FFC_metric}
\vspace{-5pt}
Intuitively, a foveated clutter model that takes into account target search should score very low when the target is 
in the fovea (near zero), and very high when the target is in the periphery. Thus, an observer should find a target without difficulty, achieving a 
near perfect hit rate in the fovea, yet the observer should have a lower hit rate in the periphery given crowding effects. 
Note that in the periphery, not only should it be harder to detect a target, but it is also likely to confuse the target with another object or region
affine in shape, size, texture and/or pixel value (false alarms).
Under this assumption, we wish to modulate a clutter score (Feature Congestion) 
by a multiplicative factor, given the target and fixation location. 
We call this multiplicative term: the PIFC coefficient,
which is defined over a $6\deg\times 6\deg$ ROI around the location of target $t$. The target itself was removed when processing the clutter maps since
it indirectly contributes to the ROI clutter score~\cite{asher2013regional}.
The PIFC aims at quantifying the information loss around the target region due to peripheral processing.

To compute the PIFC, we use the before mentioned ROI, and calculate a mean difference from 
the foveated clutter map with respect to the original non-foveated clutter map. 
If the target is foveated, there should be little to no difference between a foveated map and the original map, thus setting the PIFC coefficient value to near zero.
However, as the target is farther away from the fovea, the PIFC coefficient should be higher given pooling effects in the periphery. 
To create a foveated map, we use Feature Congestion and apply max pooling on each pooling region after the peripheral architecture has been
stacked on top of the Feature Congestion map. Note that the FFC map values will depend on the fixation location as shown in Fig.~\ref{fig:Foveated_FC}. 
The PIFC map is the result of subtracting the foveated map from the unfoveated map in the ROI, and the score is a mean distance value between these two 
maps (we use L1-norm, L2-norm or KL-divergence). Computational details can be seen in Algorithm~\ref{alg:PIFC}. Thus, we can resume our model in Eq.~\ref{main_eq}:

\begin{equation}
\label{main_eq}
 \text{FFC}_I^{f,t} = \text{FC}_I \times \text{PIFC}_{ROI(t)}^f 
\end{equation}

where $\text{FC}_I$ is the Feature Congestion score~\cite{rosenholtz2007measuring} of image $I$ which is computed by the mean of the Feature Congestion
map $R_{FC}$, and $\text{FFC}_I^{f,t}$ is the Foveated Feature Congestion score of the image $I$, depending on the point of fixation $f$ and the location of the target $t$.

\subsection{Foveated Feature Congestion Evaluation}
\label{Sec:FFC_evaluation}
\begin{figure}[t]
\centering
\subfigure[Feature Congestion with image ID's]{
    \includegraphics[scale=0.45,clip=true,draft=false,]{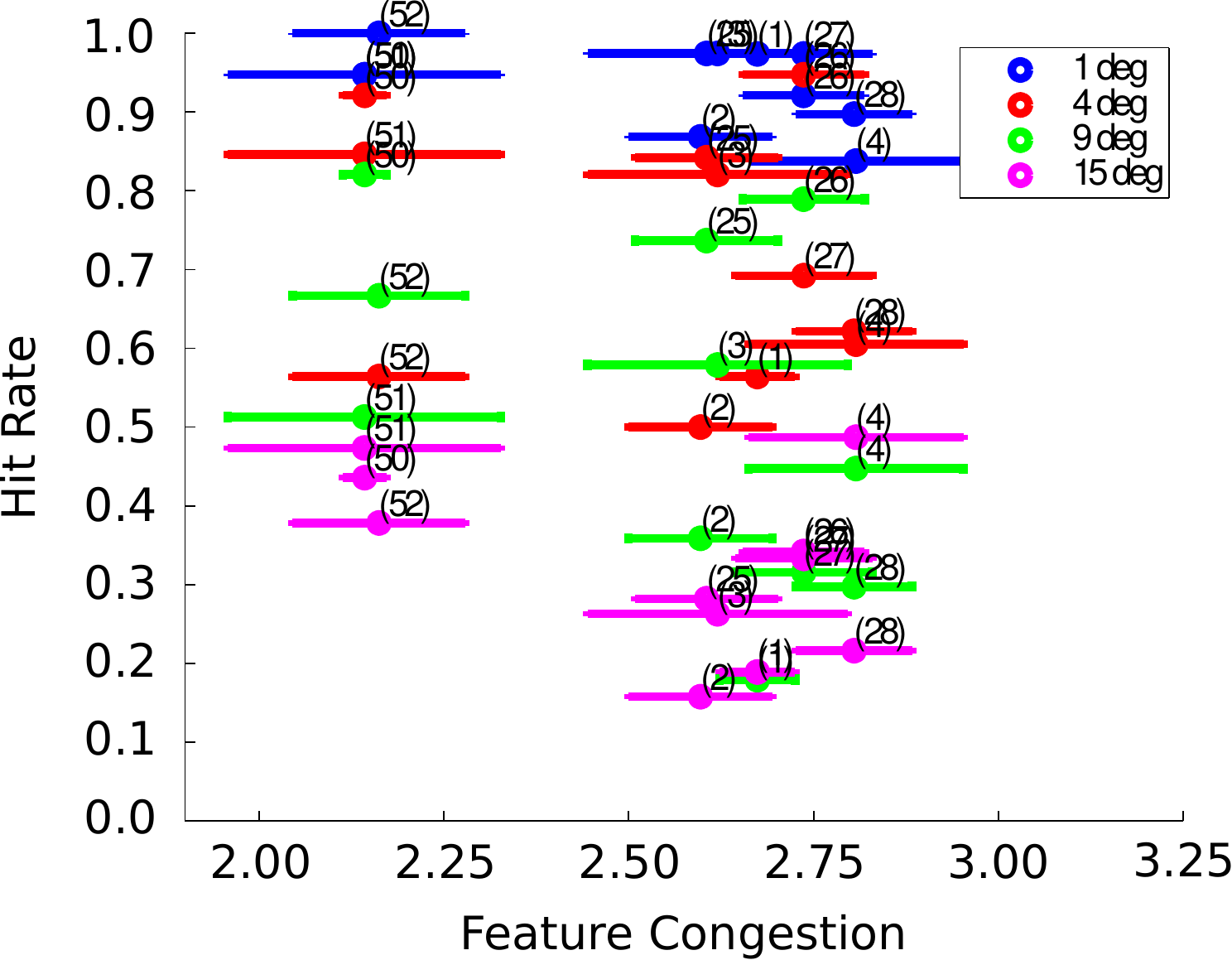}
    \label{fig:HR_FC_plot}
}
\subfigure[Foveated Feature Congestion with image ID's]{
    \includegraphics[scale=0.45,clip=true,draft=false,]{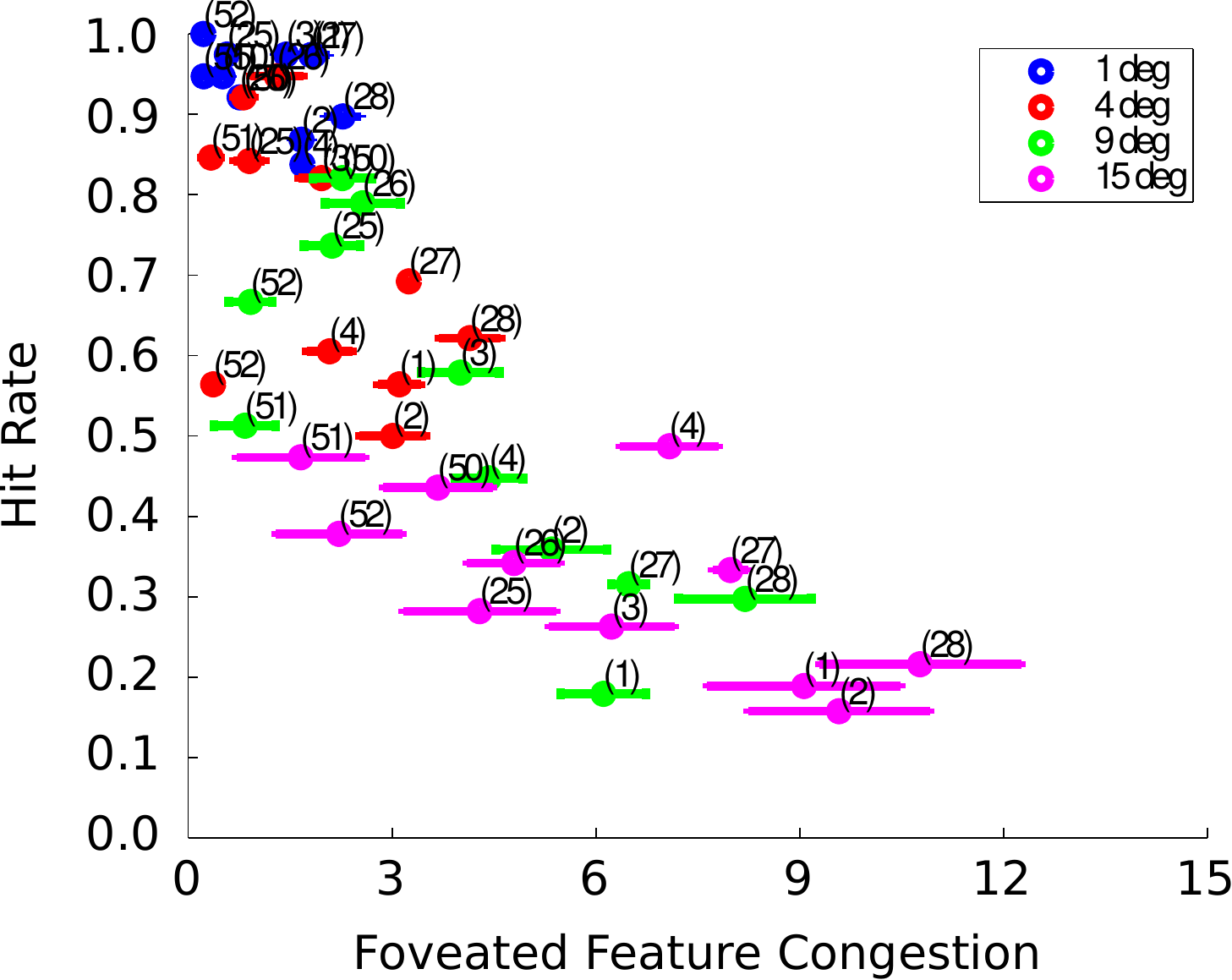}
    \label{fig:HR_FFC_plot}
}
\vspace{-10pt}
\caption{Fig.~\ref{fig:HR_FC_plot} shows the current limitations of global clutter metrics when engaging in Forced Fixation Search. The same image under different eccentricities
has the same clutter score yet possess a different hit rate. Our proposed foveated model (Fig.~\ref{fig:HR_FFC_plot}), compensates this difference through the PIFC coefficient, and modulates
each clutter score depending on fixation distance from target.
}\label{fig:HR_FC_Compare}
\end{figure}

\vspace{-5pt}
A visualization of each image and its respective Hit Rate vs Clutter Score across both foveated and unfoveated models can be 
visualized in Fig~\ref{fig:HR_FC_Compare}.
Qualitatively, it shows the importance of a PIFC weighting term to the total image clutter score when performing our forced fixation search experiment.
Futhermore, a quantitative bootstrap correlation analysis comparing classic metrics (Image, Target, ROI) against foveal metrics (FFC$_1$, FFC$_2$ and FFC$_3$) 
shows that hit rate \textit{vs} clutter scores are greater for those foveated models with a PIFC: 
Image: $(r(44)=-0.19\pm0.13,p=0.0774)$, 
Target: $(r(44)=-0.03\pm0.14,p=0.4204)$,
ROI: $(r(44)=-0.25\pm0.14,p=0.0392)$,
FFC$_1$ (L1-norm): $(r(44)=-0.82\pm0.04,p<0.0001)$, 
FFC$_2$ (L2-norm): $(r(44)=-0.79\pm0.06,p<0.0001)$,
FFC$_3$ (KL-divergence): $(r(44)=-0.82\pm0.04,p<0.0001)$.

\begin{algorithm}[t]
\small
\caption{Computation of Peripheral Integration Feature Congestion (PIFC) Coefficient}
\label{alg:PIFC}
\begin{algorithmic}[1]
\Procedure{Compute PIFC of ROI of Image $I$ on fixation $f$}{} \\
Create a Peripheral Architecture $\mathbf{A}:(N_{\theta},N_e)$\\
Offset image $I$ in Peripheral Architecture by fixation $f:(f_x,f_y)$. \\
Compute Regular Feature Congestion $(R_{FC})$ map of image $I$ \\
Set Peripheral Feature Congestion $(P_{FC}^f)\subset \mathbb{R}_{+}^2$ map to zero. \\
Copy Feature Congestion values in fovea $r_0$: $P_{FC}^f(r_0)=(R_{FC}(r_0))$ 
\For{each pooling region $r_i$ overlapping $I$, s.t. $1\leq i\leq N_{\theta}\times N_e$}
\State Get Regular Feature Congestion (FC) values in $r_i$
\State Set Peripheral FC value to max Regular FC value: $P_{FC}^f(r_i)=\max(R_{FC}(r_i))$
\EndFor \\
Crop PIFC map to ROI: $p_{FC}^f = P_{FC}^f(ROI)$ \\
Crop FC map to ROI: $r_{FC} = R_{FC}(ROI)$ \\
Choose Distance metric $D$ between $r_{FC}$ and $p_{FC}^f$ map \\
Compute Coefficient = $\text{mean}(D(r_{FC},p_{FC}^f))$ \\
\Return Coefficient
\EndProcedure
\end{algorithmic}
\end{algorithm}

Notice that there is no difference in correlations between using the L1-norm, L2-norm or KL-divergence distance for each model in terms of the correlation with 
hit rate. Table~\ref{table:param_search}(Supp. Mat.) also shows the highest correlation with a $6\times 6\deg$ ROI window across all metrics. Note that the same analysis can not be applied to false alarms, since
it is indistinguishable to separate a false alarm at $1\deg$ from $15\deg$ (the target is not present, so there is no real eccentricity away from fixation).
However as mentioned in the Methods section, fixation location for target absent trials in the experiment were placed assuming a location from its 
matching target present image. 
It is important that target present and absent fixations have the same distributions for each eccentricity.


\vspace{-10pt}
\section{Discussion}
\label{Sec:Discussion}
\begin{figure}[t]
\centering
\subfigure[FC \textit{vs} Eccentricity.]{
    \includegraphics[scale=0.295,clip=true,draft=false,]{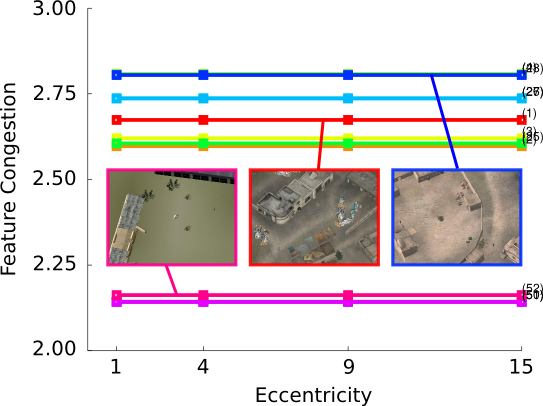}
    \label{fig:FC_plot}
}
\subfigure[PIFC (L1-norm) \textit{vs} Eccentricity.]{
    \includegraphics[scale=0.295,clip=true,draft=false,]{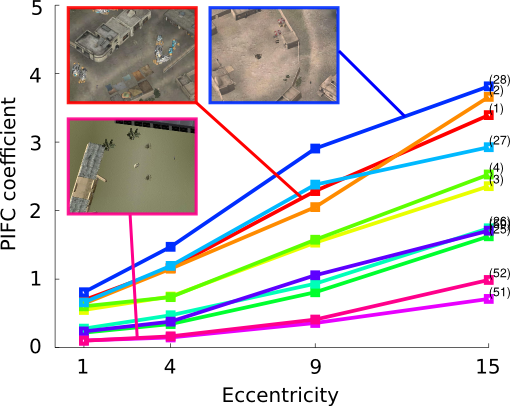}
    \label{fig:PIFC_plot}
}
\subfigure[FFC \textit{vs} Eccentricity.]{
    \includegraphics[scale=0.295,clip=true,draft=false,]{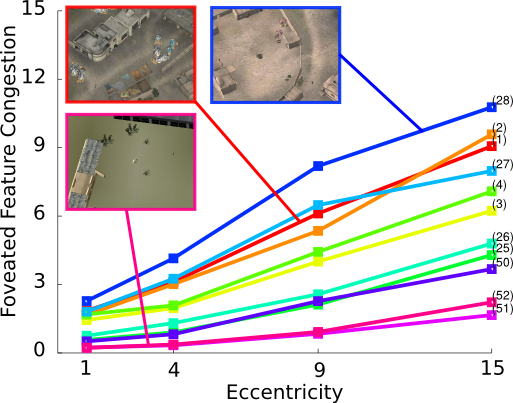}
    \label{fig:FFC_plot}
}


\vspace{-10pt}
\caption{Feature Congestion (FC) \textit{vs} Foveated Feature Congestion (FFC). In Fig.~\ref{fig:FC_plot} we see that clutter stays constant across different eccentricities
for a forced fixation task. Our FFC model (Fig.~\ref{fig:FFC_plot}) enriches the FC model, by showing how clutter increases as a function of eccentricity
through the PIFC in Fig.~\ref{fig:PIFC_plot}. 
}\label{fig:FC_vs_FFC}
\vspace{-10pt}
\end{figure}

\vspace{-10pt}
In general, images that have low Feature Congestion have less gain in PIFC coefficients as eccentricity increases. While images with high clutter
have higher gain in PIFC coefficients. Consequently, the difference of FFC between different images increases nonlinearly with eccentricity,
as observed in Fig.~\ref{fig:FC_vs_FFC}. 
This is our main contribution, as these differences in clutter score as a function of eccentricity do not exist for 
regular Feature Congestion, and these differences in scores should be able to correlate with human performance in target detection.

Our model is also different from the van der Berg~\textit{et al.}~\cite{van2009crowding} model since our peripheral architecture uses:
a biologically inspired peripheral architecture with log polar regions that provide anisotropic pooling~\cite{levi2011visual} rather than isotropic gaussian pooling 
as a linear function of eccentricity~\cite{van2009crowding}; we used region-based max pooling for each final feature map instead of pixel-based mean pooling (gaussians) per each scale (which allows for stronger
differences); this final difference also makes our model computationally more efficient running at ~700ms per image, \textit{vs} 180s per image for the 
Crowding model ($\times250$ speed up). A 
home-brewed Crowding Model applied to our forced fixation experiment resulted in a correlation of $(r(44)=-0.23\pm0.13,p=0.0469)$, equivalent to using a non 
foveated metric such as regular Feature Congestion $(r(44)=-0.19\pm0.13,p=0.0774)$.

\begin{figure}[t]
\centering
\includegraphics[scale=0.21,clip=true,draft=false,]{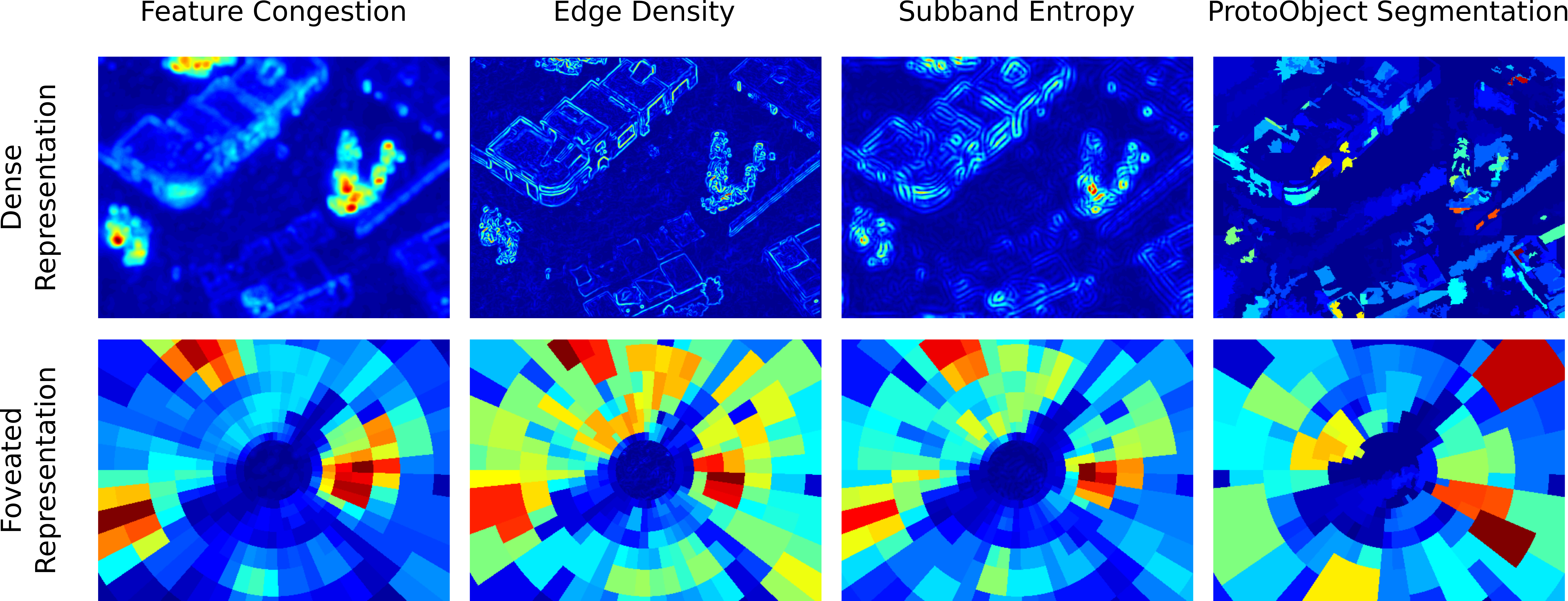}
\caption{Dense and Foveated representations of multiple models assuming a center point of fixation.
}\label{fig:Peripheral_Models_All}\vspace{-10pt}
\end{figure}

We finally extended our model to create foveated(FoV) versions of Edge Density(ED)~\cite{oliva2004identifying}, Subband 
Entropy(SE)~\cite{simoncelli1995steerable,rosenholtz2007measuring} and
ProtoObject Segmentation(PS)~\cite{yu2014modeling} showing that correlations for all foveated versions are stronger than non-foveated versions 
for the same task: $r_{ED}=-0.21$, $r_{ED+FoV}=-0.76$, $r_{SE}=-0.19$, $r_{SE+FoV}=-0.77$, $r_{PS}=-0.30$, but $r_{PS+FoV}=-0.74$. 
Note that the highest foveated correlation is FC: $r_{FC+FoV}=-0.82$, despite $r_{FC}=-0.19$ under a L1-norm loss of the PIFC. 
Feature Congestion has a dense representation, is more bio-inspired than the other models, and outperforms in the periphery. See Figure~\ref{fig:Peripheral_Models_All}.
An overview of creating dense and foveated versions for previously mentioned models can be seen in the Supp. Material.

\vspace{-10pt}
\section{Conclusion}
\vspace{-10pt}
In this paper we have introduced a peripheral architecture that shows detrimental effects of different eccentricities on target detection, that 
helps us model clutter for forced fixation experiments. We introduced a forced fixation experimental design for clutter research; we defined a biologically inspired 
peripheral architecture that pools features in V1; and we stacked the previously mentioned peripheral architecture on top of a Feature Congestion map 
to create a Foveated Feature Congestion (FFC) model -- and we extended this pipeline to other clutter models. 
We showed that the FFC model better explains loss in target detection performance as a function of eccentricity through the introduction of the
Peripheral Integration Feature Congestion (PIFC) coefficient which varies non linearly. 

\section*{Acknowledgements}

We would like to thank Miguel Lago and Aditya Jonnalagadda for useful proof-reads and revisions, as well as Mordechai Juni, N.C. Puneeth, and 
Emre Akbas for useful suggestions. This work was supported by the Institute for Collaborative 
Biotechnologies through grant 2 W911NF-09-0001 from the U.S. Army Research Office.

{\footnotesize
\bibliographystyle{ieee}
\bibliography{nips_latex}

\begin{thebibliography}{10}\itemsep=-1pt

\bibitem{achanta2010slic}
R.~Achanta, A.~Shaji, K.~Smith, A.~Lucchi, P.~Fua, and S.~S{\"u}sstrunk.
\newblock Slic superpixels.
\newblock Technical report, 2010.

\bibitem{akbas2014object}
E.~Akbas and M.~P. Eckstein.
\newblock Object detection through exploration with a foveated visual field.
\newblock {\em arXiv preprint arXiv:1408.0814}, 2014.

\bibitem{asher2013regional}
M.~F. Asher, D.~J. Tolhurst, T.~Troscianko, and I.~D. Gilchrist.
\newblock Regional effects of clutter on human target detection performance.
\newblock {\em Journal of vision}, 13(5):25--25, 2013.

\bibitem{bravo2008scale}
M.~J. Bravo and H.~Farid.
\newblock A scale invariant measure of clutter.
\newblock {\em Journal of Vision}, 8(1):23--23, 2008.

\bibitem{burt1983laplacian}
P.~J. Burt and E.~H. Adelson.
\newblock The laplacian pyramid as a compact image code.
\newblock {\em Communications, IEEE Transactions on}, 31(4):532--540, 1983.

\bibitem{Deza2016piranhas}
A.~Deza, E.~Abkas, and M.~P. Eckstein.
\newblock Piranhas toolkit: Peripheral architectures for natural, hybrid and
  artificial systems.

\bibitem{eckstein2011visual}
M.~P. Eckstein.
\newblock Visual search: A retrospective.
\newblock {\em Journal of Vision}, 11(5):14--14, 2011.

\bibitem{felzenszwalb2004efficient}
P.~F. Felzenszwalb and D.~P. Huttenlocher.
\newblock Efficient graph-based image segmentation.
\newblock {\em International Journal of Computer Vision}, 59(2):167--181, 2004.

\bibitem{freeman2011metamers}
J.~Freeman and E.~P. Simoncelli.
\newblock Metamers of the ventral stream.
\newblock {\em Nature neuroscience}, 14(9):1195--1201, 2011.

\bibitem{freeman2013functional}
J.~Freeman, C.~M. Ziemba, D.~J. Heeger, E.~P. Simoncelli, and J.~A. Movshon.
\newblock A functional and perceptual signature of the second visual area in
  primates.
\newblock {\em Nature neuroscience}, 16(7):974--981, 2013.

\bibitem{fukunaga1975estimation}
K.~Fukunaga and L.~D. Hostetler.
\newblock The estimation of the gradient of a density function, with
  applications in pattern recognition.
\newblock {\em Information Theory, IEEE Transactions on}, 21(1):32--40, 1975.

\bibitem{henderson2009influence}
J.~M. Henderson, M.~Chanceaux, and T.~J. Smith.
\newblock The influence of clutter on real-world scene search: Evidence from
  search efficiency and eye movements.
\newblock {\em Journal of Vision}, 9(1):32--32, 2009.

\bibitem{keshvari2016pooling}
S.~Keshvari and R.~Rosenholtz.
\newblock Pooling of continuous features provides a unifying account of
  crowding.
\newblock {\em Journal of Vision}, 16(39), 2016.

\bibitem{landy1991texture}
M.~S. Landy and J.~R. Bergen.
\newblock Texture segregation and orientation gradient.
\newblock {\em Vision research}, 31(4):679--691, 1991.

\bibitem{levi2011visual}
D.~M. Levi.
\newblock Visual crowding.
\newblock {\em Current Biology}, 21(18):R678--R679, 2011.

\bibitem{levinshtein2009turbopixels}
A.~Levinshtein, A.~Stere, K.~N. Kutulakos, D.~J. Fleet, S.~J. Dickinson, and
  K.~Siddiqi.
\newblock Turbopixels: Fast superpixels using geometric flows.
\newblock {\em Pattern Analysis and Machine Intelligence, IEEE Transactions
  on}, 31(12):2290--2297, 2009.

\bibitem{liu2011entropy}
M.-Y. Liu, O.~Tuzel, S.~Ramalingam, and R.~Chellappa.
\newblock Entropy rate superpixel segmentation.
\newblock In {\em Computer Vision and Pattern Recognition (CVPR), 2011 IEEE
  Conference on}, pages 2097--2104. IEEE, 2011.

\bibitem{movshon2014representation}
J.~A. Movshon and E.~P. Simoncelli.
\newblock Representation of naturalistic image structure in the primate visual
  cortex.
\newblock In {\em Cold Spring Harbor symposia on quantitative biology},
  volume~79, pages 115--122. Cold Spring Harbor Laboratory Press, 2014.

\bibitem{oliva2004identifying}
A.~Oliva, M.~L. Mack, M.~Shrestha, and A.~Peeper.
\newblock Identifying the perceptual dimensions of visual complexity of scenes.

\bibitem{portilla2000parametric}
J.~Portilla and E.~P. Simoncelli.
\newblock A parametric texture model based on joint statistics of complex
  wavelet coefficients.
\newblock {\em International Journal of Computer Vision}, 40(1):49--70, 2000.

\bibitem{rosenholtz2012summary}
R.~Rosenholtz, J.~Huang, A.~Raj, B.~J. Balas, and L.~Ilie.
\newblock A summary statistic representation in peripheral vision explains
  visual search.
\newblock {\em Journal of vision}, 12(4):14--14, 2012.

\bibitem{rosenholtz2005feature}
R.~Rosenholtz, Y.~Li, J.~Mansfield, and Z.~Jin.
\newblock Feature congestion: a measure of display clutter.
\newblock In {\em Proceedings of the SIGCHI conference on Human factors in
  computing systems}, pages 761--770. ACM, 2005.

\bibitem{rosenholtz2007measuring}
R.~Rosenholtz, Y.~Li, and L.~Nakano.
\newblock Measuring visual clutter.
\newblock {\em Journal of vision}, 7(2):17--17, 2007.

\bibitem{simoncelli1995steerable}
E.~P. Simoncelli and W.~T. Freeman.
\newblock The steerable pyramid: A flexible architecture for multi-scale
  derivative computation.
\newblock In {\em icip}, page 3444. IEEE, 1995.

\bibitem{van2009crowding}
R.~van~den Berg, F.~W. Cornelissen, and J.~B. Roerdink.
\newblock A crowding model of visual clutter.
\newblock {\em Journal of Vision}, 9(4):24--24, 2009.

\bibitem{yu2013modeling}
C.-P. Yu, W.-Y. Hua, D.~Samaras, and G.~Zelinsky.
\newblock Modeling clutter perception using parametric proto-object
  partitioning.
\newblock In {\em Advances in Neural Information Processing Systems}, pages
  118--126, 2013.

\bibitem{yu2014modeling}
C.-P. Yu, D.~Samaras, and G.~J. Zelinsky.
\newblock Modeling visual clutter perception using proto-object segmentation.
\newblock {\em Journal of vision}, 14(7):4--4, 2014.

\end{thebibliography}
}

\newpage

\section*{Supplementary Material}
\subsection*{PIFC maps}
\begin{figure}[!h]
\centering
\includegraphics[scale=0.16,clip=true,draft=false,]{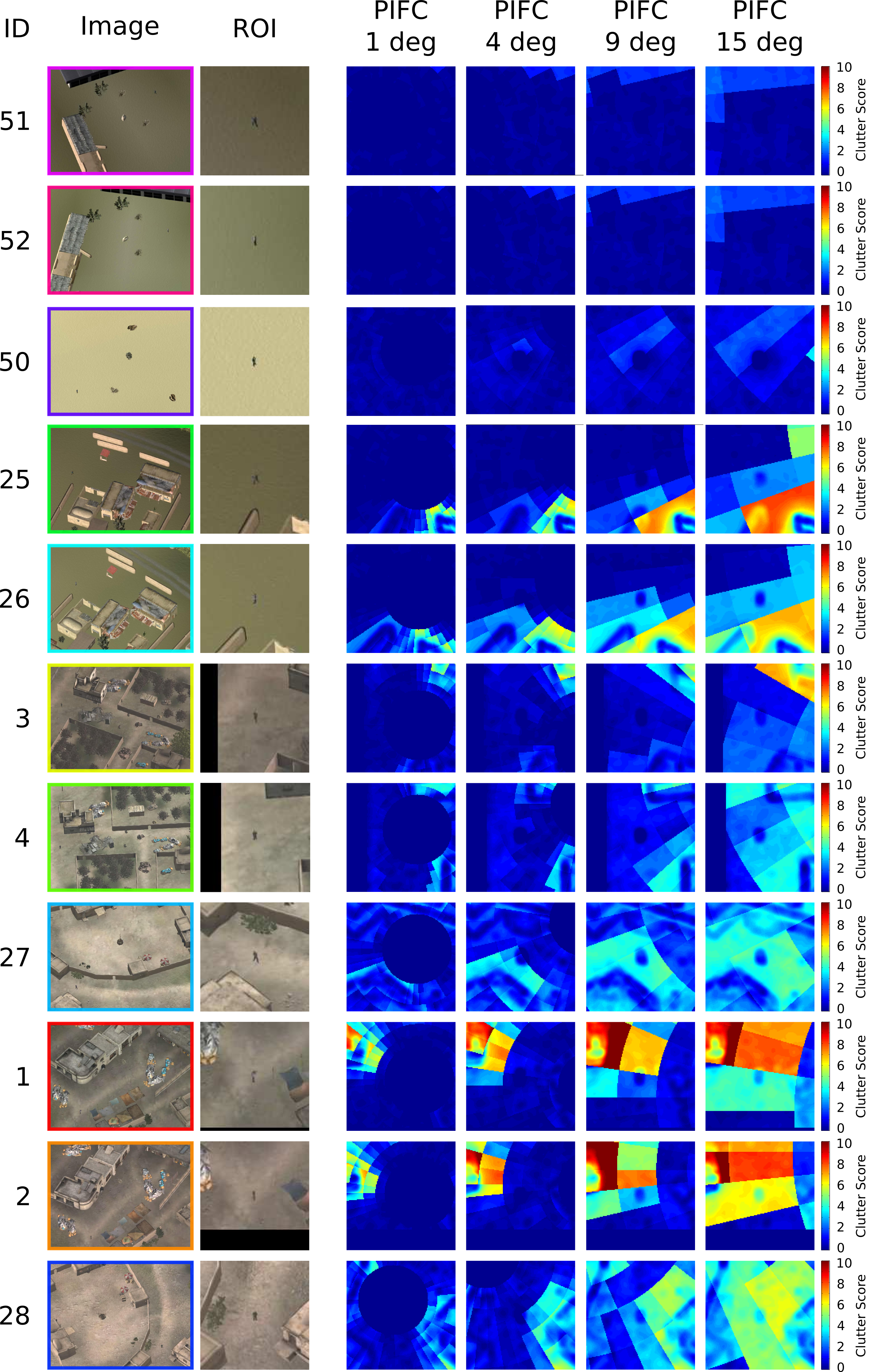}
\vspace{-10pt}
\caption{PIFC maps across the images used for our analysis ranked from least (top) to highest (bottom) FFC clutter as shown in Fig.6. Notice 
how the clutter scores (heatmap values) in the PIFC increase as a function of eccentricity and is contingent on the amount of clutter in the ROI.
}\label{fig:PIFC_Supplement}
\end{figure}

\subsection*{Beyond Foveated Feature Congestion}

We extended other clutter models to their respective peripheral versions. Since the other models: Edge Density, Subband Entropy and ProtoObject Segmentation
have not been designed to produce an intermediate step with a dense clutter pixel-wise representation (unlike Feature Congestion~\ref{fig:FC_pipeline}),
it is hard to find respective optimal dense clutter representations without losing the essence of each model. For Edge Density, we compute
the magnitude of the image gradient after grayscale conversion. For Subband Entropy, we decided to keep all the respective subbands, as
the model proposes as well as the coefficients that are used to compute a weighted sum over the entropies. In other words, our dense version of Subband Entropy
is more of a dense ``Subband Energy'' term, since computing Entropy over a vector of a small $N\times K$ vector space of $N=3$ scales and $K=4$ orientations 
produced very little room for variation.
Finally dense ProtoObject Segmentation was computed by following the intuition of final 
number of superpixels over inital number of superpixels, but since this is not applicable 
at a pixel wise level, we decided to compute multiple ProtoObject Segmentations with different regularizer and superpixel radius parameters, and averaged all superpixel
segmentation ratios -- where every map was dense at a superpixel level, and each superpixel score was the initial number of pixels 
over the final number of initial number of pixels that belong to that superpixel after the meanshift merging stage in HSV color space. 


We believe that future work can be tailored towards improving dense versions of each clutter model, as well as creating new dense clutter models, that can
easily be stacked with a peripheral architecture.

\begin{table}[h]
\small
\begin{tabular}{|c|c|c|c|c|c|}
 \hline
 \multicolumn{6}{|c|}{Foveated Feature Congestion \textit{vs} Hit Rate correlation} \\
 \hline
Distance &  $4\deg$ & $6\deg$ & $8\deg$ & $10\deg$ & $12\deg$ \\
 \hline
L1-norm       & $-0.80\pm0.04$ & $\mathbf{-0.82\pm0.04}$ & $-0.81\pm0.05$ & $-0.79\pm0.05$ & $-0.76\pm0.06$ \\
L2-norm       & $\mathbf{-0.79\pm0.05}$ & $\mathbf{-0.79\pm0.06}$ & $-0.77\pm0.06$ & $-0.75\pm0.07$ & $-0.71\pm0.07$  \\
KL-divergence & $-0.80\pm0.04$ & $\mathbf{-0.82\pm0.04}$ & $\mathbf{-0.82\pm0.04}$ & $-0.81\pm0.05$ & $-0.77\pm0.06$ \\
\hline
\hline
 \multicolumn{6}{|c|}{Foveated Edge Density \textit{vs} Hit Rate correlation} \\
 \hline
Distance &  $4\deg$ & $6\deg$ & $8\deg$ & $10\deg$ & $12\deg$ \\
 \hline
L1-norm       & $\mathbf{-0.76\pm0.06}$ & $-0.73\pm0.07$ & $-0.69\pm0.08$ & $-0.65\pm0.09$ & $-0.59\pm0.09$ \\
L2-norm       & $\mathbf{-0.72\pm0.07}$ & $-0.66\pm0.08$ & $-0.62\pm0.09$ & $-0.56\pm0.10$ & $-0.50\pm0.11$ \\
KL-divergence & $\mathbf{-0.76\pm0.06}$ & $\mathbf{-0.76\pm0.06}$ & $-0.73\pm0.07$ & $-0.69\pm0.08$ & $-0.63\pm0.08$  \\
\hline
\hline
 \multicolumn{6}{|c|}{Foveated Subband Entropy \textit{vs} Hit Rate correlation} \\
 \hline
Distance &  $4\deg$ & $6\deg$ & $8\deg$ & $10\deg$ & $12\deg$ \\
 \hline
L1-norm       & $-0.75\pm0.04$ & $\mathbf{-0.77\pm0.04}$ & $\mathbf{-0.77\pm0.05}$ & $-0.76\pm0.05$ & $-0.73\pm0.06$ \\
L2-norm       & $-0.74\pm0.05$ & $\mathbf{-0.76\pm0.05}$ & $\mathbf{-0.76\pm0.05}$ & $-0.75\pm0.06$ & $-0.71\pm0.06$ \\
KL-divergence & $-0.79\pm0.04$ & $-0.83\pm0.04$ & $\mathbf{-0.84\pm0.04}$ & $-0.83\pm0.04$ & $-0.80\pm0.05$ \\
\hline
\hline
\multicolumn{6}{|c|}{Foveated ProtoObject Segmentation \textit{vs} Hit Rate correlation} \\
\hline
Distance &  $4\deg$ & $6\deg$ & $8\deg$ & $10\deg$ & $12\deg$ \\
\hline
L1-norm       & $-0.70\pm0.06$ & $\mathbf{-0.74\pm0.06}$ & $\mathbf{-0.74\pm0.06}$ & $-0.72\pm0.06$ & $-0.66\pm0.07$ \\
L2-norm       & $-0.74\pm0.04$ & $\mathbf{-0.76\pm0.05}$ & $\mathbf{-0.76\pm0.05}$ & $-0.76\pm0.06$ & $-0.72\pm0.06$ \\
KL-divergence & $-0.66\pm0.06$ & $\mathbf{-0.71\pm0.05}$ & $-0.68\pm0.06$ & $-0.61\pm0.07$ & $-0.54\pm0.08$ \\
\hline
\end{tabular}
\caption{Foveated Clutter Models distance and ROI window length $(\deg)$ search.}
\label{table:param_search}
\end{table}



\end{document}